\documentclass[graybox, natbib, nosecnum]{svmult}
\bibpunct{(}{)}{;}{a}{}{,} %

\usepackage{mathptmx}       %
\usepackage{helvet}         %
\usepackage{courier}        %
\usepackage{type1cm}        %

\usepackage{makeidx}         %
\usepackage{graphicx}        %
\usepackage{multicol}        %
\usepackage[bottom]{footmisc}%
\usepackage[normalem]{ulem}	%
\usepackage{hyperref}  %
\usepackage{soul}   %
\usepackage[absolute]{textpos}
\definecolor{somegray}{rgb}{0.5, 0.5, 0.5}
\newcommand{\darkgrayed}[1]{\textcolor{somegray}{#1}}

\begin{document}

\begin{textblock}{20}(2,1)
	\noindent\large \darkgrayed{This paper has been accepted for publication in the
		Springer Encyclopedia of Robotics, 2019. \copyright Springer}
\end{textblock}

\title*{Visual-Inertial Odometry of Aerial Robots}

\author{Davide Scaramuzza and Zichao Zhang}
\institute{Davide Scaramuzza and Zichao Zhang \at Dep. of Neuroinformatics, ETH Zurich and University of Zurich\\
Dep. of Informatics, University of Zurich, \\
Zurich, Switzerland\\ 
\email{sdavide@ifi.uzh.ch}\\
\email{zzhang@ifi.uzh.ch}}
\maketitle
\section{Synonyms}
Visual-inertial state estimation, inertial-aided vision-based state estimation, Visual-Inertial Simultaneous Localization and Mapping (VISLAM), camera tracking, ego-motion estimation.

\section{Definitions}
Visual-Inertial odometry (VIO) is the process of estimating the state (pose and velocity) of an agent (e.g., an aerial robot) by using only the input of one or more cameras plus one or more Inertial Measurement Units (IMUs) attached to it. VIO is the only viable alternative to GPS and lidar-based odometry to achieve accurate state estimation. Since both cameras and IMUs are very cheap, these sensor types are ubiquitous in all today's aerial robots.

\section{Overview}\label{Overview}
Cameras and IMUs are complementary sensor types.
A camera accumulates the photons during the exposure time to get a 2D image.
Therefore they are precise during slow motion 
and provide rich information, which is useful for other perception 
tasks, such as place recognition.
However, they have limited output rate ($\sim$100Hz), suffer from scale 
ambiguity in a monocular setup, and are not robust to scenes 
characterized by low 
texture, high speed motions (due to motion blur) and High Dynamic Range 
(HDR) (which may cause over- or under-exposure of the image). 
By contrast, an IMU is a proprioceptive sensor measuring the angular velocity and the external acceleration acting upon it.
An IMU is scene-independent, which renders it unaffected by the aforementioned difficulties for cameras.
Thus, it is the ideal complement to cameras to achieve robustness in low texture, high speed, and HDR scenarios.
Additionally, an IMU has high output rate ($\sim$1,000Hz). 
However, it suffers from poor signal-noise ratio at low accelerations and low angular velocities.
Due to the presence of sensor biases, the motion estimated from an IMU alone tends to accumulate drift quickly.
Therefore, a combination of both cameras and IMUs can provide accurate and robust state estimation in different situations.

\begin{figure}[t]
 \includegraphics[width=0.45\linewidth]{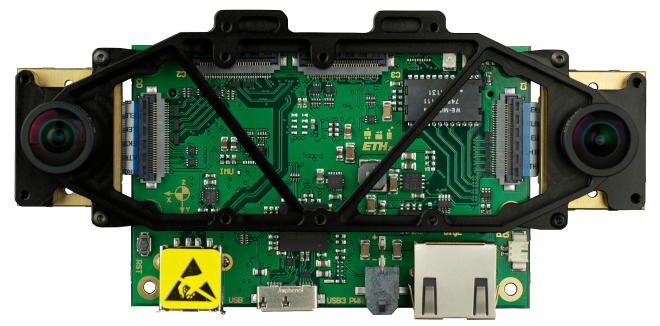}
 \includegraphics[width=0.5\linewidth]{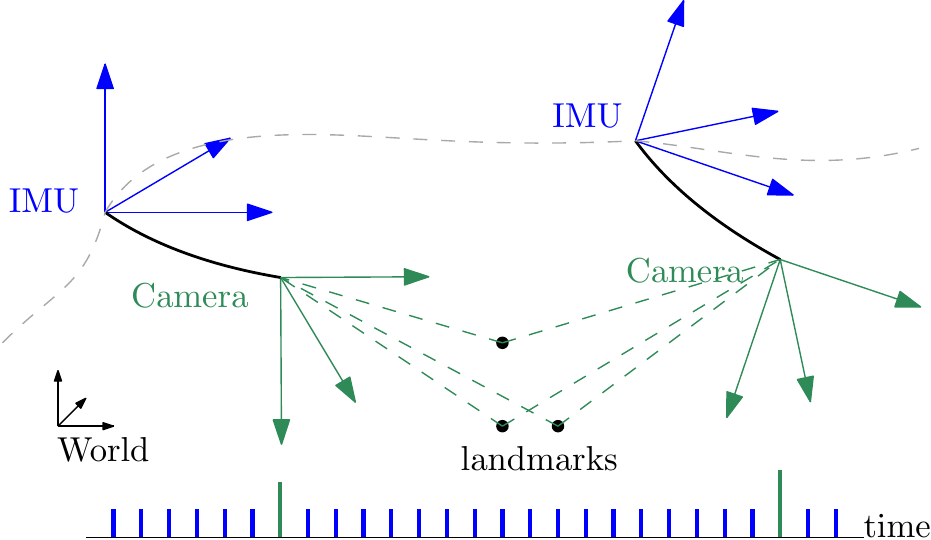}
 \caption{Illustration of visual-inertial sensors and measurements.
 Left: Stereo VI-sensor from \citep{Nikolic14icra}.
 Right: Camera (green) and IMU  (blue) measurements along a trajectory.}\label{fig:vio}
\end{figure}
The typical VIO configuration is illustrated in Fig.~\ref{fig:vio}.
The camera(s) and IMU(s) are rigidly attached, and the sensor suite outputs visual and inertial measurements at different rates.
In VIO, the environment  is represented as a set of 3D landmarks ${}_{W}\mathbf{p}$ that are projected by the camera to 2D image coordinates $\mathbf{u}$:
\begin{equation}
\mathbf{u} = project(\mathbf{T}_{CW} \cdot {}_{W}\mathbf{p}).
\label{eq:cam_meas}
\end{equation}
The IMU measures the angular velocity $\omega$ and the external acceleration $\mathbf{a}$:
\begin{equation}
\omega = {}_{I}\omega + \mathbf{b}_g + \mathbf{n}_g, \quad 
\mathbf{a} = \mathbf{R}_{IW}({}_{W}\mathbf{a} - {}_{W}\mathbf{g}) + \mathbf{b}_a + \mathbf{n}_a,
\label{eq:imu_meas}
\end{equation}
where ${}_{I}\omega$ is the angular velocity of the IMU expressed in the IMU frame, ${}_{W}\mathbf{a}$ the acceleration of the IMU in the world frame, and ${}_{W}\mathbf{g}$ the gravity in the world frame. $\mathbf{b}$ and $\mathbf{n}$ are the biases and additive noises respectively (see \citep{Furgale13iros} for details). It is also worth mentioning that for low-cost IMUs, the above model can be over simplified, and additional errors from scale factors and axis misalignment sometimes also need to be considered~\citep{rehder2016ICRA}.

VIO is the process of estimating the state of the sensor suite using the camera and IMU measurements (\ref{eq:cam_meas}) and (\ref{eq:imu_meas}).
Typically, the quantities to estimate are $N$ \emph{states} at different times $\{t_i\}_{i=1}^{N}$
\begin{equation}
\mathbf{X}_i = [\mathbf{T}_{WI}^i,~\mathbf{v}_{WI}^i,~\mathbf{b}_a^i,~\mathbf{b}_g^i], \quad i = 1,~2,~3,~\ldots,~N
\label{eq:state}
\end{equation}
where $\mathbf{T}_{WI}^i$ is the 6-DoF pose of the IMU, $\mathbf{v}_{WI}^i$ is the velocity of the IMU, $\mathbf{b}_a^i$ and $\mathbf{b}_a^i$ are the biases of the accelerometer and gyroscope respectively.
In contrast to visual-only odometry, the velocity and the biases are essential to utilize the IMU measurements and have to be estimated in addition to the 6-DoF pose.
Specifically, biases are necessary for computing the actual sensor angular velocity and acceleration from the raw measurements (\ref{eq:imu_meas}), and velocity is needed for integrating acceleration to get position.

VIO can utilize multiple cameras and IMUs; however, the minimum number of cameras and IMUs that is sufficient to perform VIO is one.
Indeed, a single moving camera allows us to measure the geometry of the 3D scene and the 
camera motion up to an unknown metric scale:  the projection function in (\ref{eq:cam_meas}) satisfies $project(\mathbf{p}) = project(s\cdot\mathbf{p})$ for an arbitrary scalar $s$ and an arbitrary point $\mathbf{p}$;
a single IMU, instead, renders metric scale and gravity observable (due to the presence of gravity in (\ref{eq:imu_meas}))~\citep{Martinelli13fntrob}.

\begin{figure}[t]
\centering
 \includegraphics[width=1.0\linewidth]{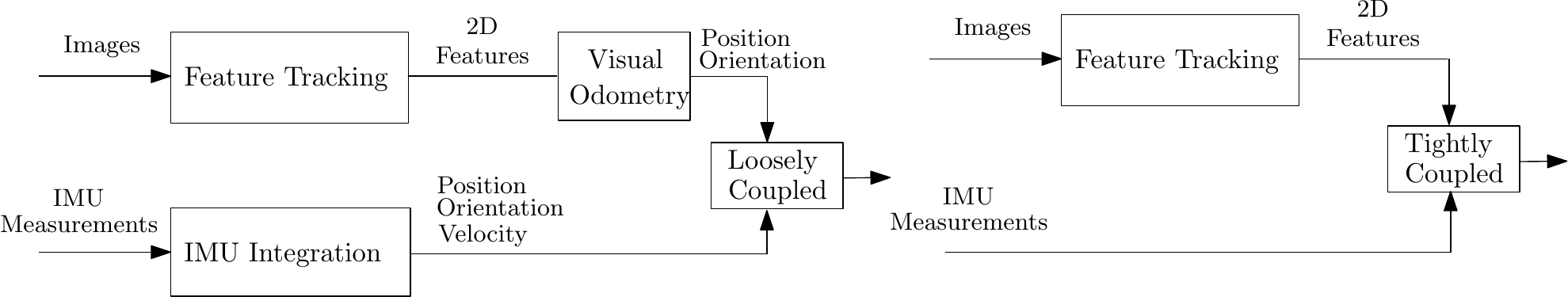}
 \caption{Comparison of loosely (left) and tightly coupled (right) paradigms for VIO.}\label{fig:coupled}
\end{figure}
Depending on the specific information used to fuse visual and inertial measurements, VIO approaches can be categorized into two paradigms: \emph{loosely coupled} and \emph{tightly coupled}~\citep{Corke07ijrr}.
Conceptually, loosely coupled methods process visual and inertial measurements separately by computing two independent motion estimates that are then fused to get the final output.
By contrast, tightly coupled methods compute the final output directly from the raw camera and IMU measurements, e.g., the tracked 2D features, angular velocities, and linear accelerations (\ref{eq:imu_meas}). The difference between these two approaches is conceptually illustrated in Fig.~\ref{fig:coupled}.

Tightly coupled approaches are more accurate than the loosely coupled ones.
First, using IMU integration to predict the 2D feature locations in the next frame can be used to facilitate feature tracking.
Second, loosely-coupled approaches do not consider the visual and
inertial information coupling, making them incapable of correcting drift in the vision-only estimator.

\section{Key Research Findings}\label{keyresearchfindings}

\subsection{The Three Major VIO Paradigms}

Existing VIO approaches can be categorized by the number of camera-poses involved in the estimation, which is highly correlated with the computational demand and accuracy.
Full smoothers (or batch nonlinear least-squares algorithms) estimate the complete history of states, fixed-lag smoothers
(or sliding window estimators) consider a window of the latest states, and filtering methods only estimate the latest state.
Full smoothers, by keeping the whole pose history, allow re-linearization when the estimate is updated.
By contrast, sliding window estimators and filters marginalize older states (which locks the linearization error permanently) and are therefore less accurate but more efficient.
Early research on VIO focuses on filters because of their efficiency. 
Recently, the research focus has shifted to fixed-lag/full smoothers due to their superior accuracy and the availability of more powerful computers.
An up to date review on visual-inertial navigation can be found in~\citep{Huang2019ICRA}.

Note that there are different criteria that can be used to characterize VIO algorithms.
VIO approaches can use different representations of the uncertainty for the measurements and the Gaussian priors:
the Extended Kalman Filter (EKF) represents the uncertainty using a covariance matrix;
instead, information filters and smoothers resort to the information matrix (the inverse of the covariance) or the square-root of the information matrix \citep{Kaess12ijrr,Wu15rss}.
The number of times in which the measurement model is linearized is also an important criterion.
While a standard EKF (in contrast to the iterated EKF) processes a measurement only once, a smoothing approach allows linearizing multiple times.

While the terminology is vast, the underlying algorithms are tightly related. For instance, it can be shown that the iterated Extended Kalman filter equations are equivalent to the Gauss-Newton algorithm, commonly used for smoothing \citep{Bell93ac}.

\subsubsection{Filtering}
Filtering algorithms enable efficient estimation by restricting the inference process to the latest state of the system.
Classic approaches estimate both the poses and landmarks, and the complexity of the filter (e.g., the Extended Kalman Filter) grows quadratically in the number of estimated landmarks.
Therefore, a small number of landmarks are typically tracked to allow real-time operation \citep{Davison07pami,Blosch15iros,Jones11ijrr}. 

An alternative is to adopt a “structureless” approach where landmark positions are marginalized out of
the state vector (see, for instance, the Multi-State Constraint Kalman filter (MSCKF)~\citep{Mourikis07icra}).
A drawback of the structureless filter is that the processing of landmark measurements needs to
be delayed until all measurements of a landmark are obtained~\citep{Mourikis07icra}.
This hinders accuracy as the filter cannot use all current visual information.

There are two major error sources for filtering approaches.
First, filters absorb the information of the older states into the estimation of the latest state and drop the older states permanently.
Therefore, linearization error and erroneous outlier measurements (see \citep{Tsotsos15icra}) are locked in the filter state.
Second, linearization error renders filters inconsistent.
Generally, the VIO problem has four unobservable directions: the global position and the orientation around the gravity direction (yaw)~\citep{Martinelli13fntrob,Kottas2012iser}. 
In~\citep{Kottas2012iser} it is shown that linearization at the wrong estimate adds spurious information in unobservable directions.
To address this problem, the first-estimates jacobian approach~\citep{Huang08iser} is often adopted to ensures that a state is not updated with different linearization points, which is a source of inconsistency.

\subsubsection{Fixed-lag Smoothing}
Fixed-lag smoothers estimate the states that fall within a given time window, while marginalizing out older states~\citep{Mourikis08wvlmp,Sibley10jfr,DongSi11icra,Leutenegger15ijrr}. 
For VIO, which is highly nonlinear, fixed-lag smoothing approaches are generally more accurate than filtering, since they relinearize part of the past measurements as the estimate is updated~\citep{Maybeck79}.
Moreover, these approaches are more robust to outliers by explicit outlier rejection after the optimization or using robust cost functions \citep{Hartley03book}.
However, since fixed-lag smoothers still resort to marginalization, they, similar to filters, suffer from inconsistency and linearization errors \citep{Hesch14ijrr,DongSi11icra,Huang11iros}.

Fixed-lag smoothers are more computationally expensive than filters, since multiple states instead of the latest one are estimated.
In addition, the marginalization of the states outside the estimation window can lead to dense Gaussian priors, which hinders efficient matrix operations.
For this reason, it has been proposed to drop certain measurements instead of marginalizing them to maintain the sparsity of the problem~\citep{Leutenegger15ijrr}.

\subsubsection{Full Smoothing}
Full smoothing methods estimate the entire history of the states by solving a large nonlinear optimization problem~\citep{Jung01cvpr,Sterlow04ijrr,Bryson09icra,Indelman13ras,PatronPerez15ijcv}. 
Full smoothing guarantees the highest accuracy, since it can update the linearization point of the complete state history as the estimate evolves.
However, because the complexity of the optimization problem is approximately cubic with respect to the dimension of the states, real-time operation quickly becomes infeasible as the trajectory and the map grow over time. 
Common practice (also widely used in fixed-lag smoothers) is to only keep selected keyframes~\citep{Leutenegger15ijrr, Qin17arxiv, Strasdat10icra, Nerurkar14icra} 
and/or run the optimization in a parallel tracking and mapping architecture~\citep{Mourikis08wvlmp,Klein09ismar}.
A breakthrough has been the development of incremental smoothing techniques (iSAM~\citep{Kaess08tro}, iSAM2~\citep{Kaess12ijrr}).
They leverage the expressiveness of factor graphs to maintain sparsity and to identify and update
only the typically small subset of variables affected by a new measurement.
VIO using the incremental smoothing framework has been demonstrated in \citep{Forster17troOnmanifold}.

The different update rates of cameras and IMUs bring additional difficulties for both full-smoothing and fixed-lag smoothing approaches.
Filtering methods usually use IMUs for the process model and cameras for the measurement model, and, thus, they handle the different rates of IMUs and cameras naturally. 
In smoothing approaches, however, it is infeasible for real-time applications to add a state at every IMU measurement, since the problem complexity grows with the dimension of the states.
Therefore, the IMU measurements are typically integrated between frames to form relative motion constraints.
This requires the integration to be repeated when the state estimate changes (i.e., after each optimization iteration).
~\citep{Lupton12tro} show that this repeated integration can be avoided by a reparametrization of the relative motion constraints.
Such reparametrization is called IMU preintegration. ~\citep{Forster17troOnmanifold} builds upon ~\citep{Lupton12tro} and bring the theory of IMU preintegration to maturity by properly addressing the manifold structure of the rotation group SO(3). 

\subsection{Camera-IMU Calibration}
The knowledge about the spatial transformations and temporal offsets between camera(s) and IMU(s) is crucial to obtain good performance from VIO.
Off-line spatial calibration of cameras and IMUs is a well-studied problem and can be solved using both filters \citep{Kelly11ijrr} and batch optimization \citep{Furgale13iros}.
For on-line self calibration, state-of-the-art VIO algorithms often include the unknown spatial transformation (between cameras and IMUs) in the states and estimate it together with the motion of the sensor suite \citep{Leutenegger15ijrr,Li13iros}.
If the visual-inertial sensor suite is not hardware synchronized as in \citep{Nikolic14icra}, the temporal offset of the cameras and IMUs also needs to be estimated.
While different approaches have been proposed for off-line calibration \citep{Kelly14iser,Furgale13iros}, very few works have been done for on-line processing \citep{Li13iros, Qin2018iros}.

A significant contribution to the community is the open source calibration toolbox \emph{Kalibr} \citep{Furgale13iros}.
It uses a continuous representation of the trajectory instead of discrete states, which has, therefore, the ability to model the temporal offset between the cameras and IMUs.
\emph{Kalibr} is widely used for both spatial and temporal calibration of camera-IMU systems.

\section{Examples of Applications}\label{Applications}
Pioneering work on VIO-based autonomous navigation of aerial robots was done in the context of the European project sFly (2009-2012)~\citep{Bloesch10icra,Weiss13jfr,Lynen13iros,Forster13irosCollaborative,Meier12ar,Scaramuzza14ram}.

Nowadays, there are several open-source VIO software
packages available that have explicitly been developed for and successfully deployed on aerial robots. 
Most of these pipelines are monocular, since a single camera and an IMU are the minimal
sensor suite necessary for reliable state estimation; the monocular setup is a convenient choice for flying robots due to its low weight and power
consumption, with respect to other sensor configurations, such as stereo or multi-camera systems. The open source monocular VIO software packages currently available are:
\begin{itemize}

 \item MSCKF~\citep{Mourikis07icra} - The Multi-State Constraint Kalman Filter (MSCKF) forms the basis
of many modern, proprietary VIO systems (such as Google ARCore, the former Google Tango), but until recently
no sufficient, publicly available implementation existed. The
original MSCKF algorithm in~\citep{Mourikis07icra} proposed a measurement
model that expressed the geometric constraints between all
of the camera poses that observed a particular image feature,
without the need to maintain an estimate of the 3D feature
position in the state. The extended Kalman filter backend
in~\citep{Zhu17cvpr} implements this formulation of the MSCKF for
event-based camera inputs, but has been adapted to feature
tracks from standard cameras. The code is publicly available at: \url{https://github.com/daniilidis-group/msckf_mono}.

 \item OKVIS~\citep{Leutenegger13rss,Leutenegger15ijrr} - Open Keyframe-based Visual-Inertial SLAM (OKVIS) utilizes non-linear optimization on a sliding window of
keyframe poses. The cost function is formulated with a
combination of weighted reprojection errors for visual landmarks
and weighted inertial error terms. The frontend uses a
multi-scale Harris corner detector~\citep{Harris88} to find features, and
then computes BRISK descriptors~\citep{Leutenegger11iccv} on them in order to
perform data association between frames. Keyframes older
than the sliding window are marginalized out of the states
being estimated. OKVIS uses Google’s Ceres solver~\citep{ceres-solver}
to perform non-linear optimization. It should be noted that
OKVIS is not optimized for monocular VIO, and in~\citep{Leutenegger15ijrr} it
shows superior performance using a stereo configuration. The
software is available in a ROS-compatible package at: \url{https://github.com/ethz-asl/okvis_ros}.

 \item ROVIO~\citep{Blosch15iros} - Robust Visual Inertial Odometry (ROVIO) is a visual-inertial
state estimator based on an extended Kalman Filter
(EKF), which proposed several novelties. In addition to
FAST corner features~\citep{Rosten10pami}, whose 3D positions are parameterized
with robot-centric bearing vectors and distances,
multi-level patches are extracted from the image stream
around these features. The patch features are tracked, warped
based on IMU-predicted motion, and the photometric errors
are used in the update step as innovation terms. Unlike
OKVIS, ROVIO was developed as a monocular VIO
pipeline. The pipeline is available as an opensource
software package at: \url{https://github.com/ethz-asl/rovio}.

 \item VINS-Mono - VINS-Mono~\citep{Qin17arxiv} is a non-linear optimization-based sliding
window estimator, tracking robust corner features~\citep{Shi94cvpr},
similar to OKVIS. However, VINS-Mono introduces several new features to this class of estimation framework. The
authors propose a loosely-coupled sensor fusion initialization
procedure to bootstrap the estimator from arbitrary initial
states. IMU measurements are pre-integrated before being
used in the optimization, and a tightly-coupled procedure for
relocalization is proposed. VINS-Mono additionally features
modules to perform 4DoF pose graph optimization and loop
closure. The software is available in both a ROS compatible
PC version and an iOS implementation for state
estimation on mobile devices. VINS-Mono is available at: \url{https://github.com/HKUST-Aerial-Robotics/VINS-Mono}.

 \item SVO+MSF - Multi-Sensor Fusion (MSF)~\citep{Lynen13iros} is a general EKF framework
for fusing data from different sensors in a state
estimate. Semi-Direct Visual Odometry (SVO)~\citep{Forster14icra,Forster17troSVO} is a
computationally lightweight visual odometry algorithm that
aligns images by tracking FAST corner features and edgelets and
minimizing the photometric error of patches around them.
This sparse alignment is then jointly optimized with the scene
structure by minimizing the reprojection error of the features
in a nonlinear least-squares optimization. The pose estimated
from the vision-only SVO is provided to MSF as the output
of a generic pose sensor, where it is then fused in a loosely-coupled manner with the IMU
data, as proposed in~\citep{Faessler16jfr}. Both MSF and SVO are publicly
available, and communicate through a ROS interface. 
SVO 2.0 is available at: \url{https://github.com/uzh-rpg/rpg_svo_example}. 
MSF is available at: \url{https://github.com/ethz-asl/ethzasl_msf}.

 \item SVO+GTSAM~\citep{Forster17troOnmanifold} - The same visual odometry frontend as in the SVO+MSF
system has also been paired with a full-smoothing backend
performing online factor graph optimization using
iSAM2~\citep{Kaess12ijrr}. In~\citep{Forster17troOnmanifold}, the authors present results using this
integrated system and propose the use of pre-integrated IMU
factors in the pose graph optimization. Both components
of this approach, SVO and the GTSAM 4.0 optimization
toolbox~\citep{Dellaert12tr}, are publicly available. SVO 2.0 is available at: \url{https://github.com/uzh-rpg/rpg_svo_example}. GTSAM is available at: \url{https://bitbucket.org/gtborg/gtsam/}.
\end{itemize}

A benchmark comparison of all the aforementioned open-source monocular VIO pipelines on common flying robot hardware  (Odroid, Up Board, and Intel NUC) has been recently published~\citep{Delmerico18icra}.
The evaluation considers the pose estimation accuracy, per-frame processing time, and CPU and memory load while processing the EuRoC Micro Aerial Vehicle datasets~\citep{Burri16ijrr}, which contain several 6DoF trajectories typical of flying robots.
Note that quantitatively evaluating the accuracy is a non-trivial task due to the unobservable DoFs in VIO systems, for which a tutorial can be found in~\citep{Zhang18iros}.

Very recently, a VIO pipeline combining an event-camera, a standard camera, and an IMU (called UltimateSLAM) has been published~\citep{ultimateSLAM}. 
It is shown on public datasets that this hybrid pipeline leads to an accuracy improvement of $85\%$ over standard, frame-based VIO
systems. Furthermore, it is shown that it can be used for autonomous quadrotor flight in scenarios inaccessible with traditional VIO, such as high-speed motion, low-light environments and high dynamic range scenes.

\section{Future Directions for Research}\label{Future}
New research directions are the integration of complementary sensors, such as
event cameras, and new algorithmic tools, such as deep learning.

Contrarily to standard cameras, which send entire images at fixed frame
rates, event cameras, such as the dynamic vision sensor
(DVS)~\citep{Lichtsteiner08ssc} or the asynchronous time-based image sensor
(ATIS)~\citep{Posch11ssc}, only send the local pixel-level brightness changes 
caused by movement in a scene at the time they occur.
Event cameras have four key advantages compared to standard cameras: 
a very low temporal latency (microseconds), a very high output
rate (up to 1 MHz vs 100Hz of standard cameras), a very high dynamic range (up to 140 dB vs 60 dB of standard cameras), and a very low power consumption (10mW vs 1W of standard cameras). 
These properties enable the design of a new class
of VIO and VISLAM algorithms that can operate in scenes characterized by high-speed motion~\citep{Gallego17pami,Rebecq17ijcv,ultimateSLAM} and high-dynamic range~\citep{Kim14bmvc,KimEccv16,Rebecq17ral,Rebecq17ijcv,ultimateSLAM},
where standard cameras fail. However, since the output is composed
of a sequence of asynchronous events, traditional frame-based
computer-vision algorithms are not directly applicable, so that novel algorithms must be developed to deal with these cameras. 

A robust VIO architecture should not solely exploit geometry and sensor measurement models but should also be able to exploit semantic/contextual information about the environment and application-specific priors about the motion dynamics. In this respect, the recent development of deep visual(-inertial) odometry~\citep{Costante16icra,Wang17icra,Zhou17cvpr, Clark17aaai} has shown promising initial results, especially in addressing open challenges with standard cameras, such as dealing with the aperture problem, motion blur, defocus, and low visibility scenarios. However, in terms of accuracy, end-to-end methods are still not on par with traditional methods currently.

\section{Cross-References}
Visual Simultaneous Localization and Mapping; Visual Odometry.

\bibliographystyle{spbasic}  %
\bibliography{all} %

\end{document}